\documentclass[11pt,a4paper]{article}
\usepackage[hyperref]{emnlp2020}
\usepackage{times}
\usepackage{latexsym}
\usepackage{amsmath}
\usepackage{symbol}
\usepackage{multirow}
\usepackage{xspace,mfirstuc,tabulary}
\usepackage{booktabs}
\usepackage{slashbox}
\usepackage{xcolor,colortbl}

\usepackage{graphicx}  
\usepackage{microtype}

\usepackage{colortbl}
\newcommand{\CC}[1]{\cellcolor{blue!#1}}
\usepackage{lipsum}

\usepackage{cleveref}
\crefformat{section}{\S#2#1#3}
\crefformat{subsection}{\S#2#1#3}
\crefformat{subsubsection}{\S#2#1#3}
\crefrangeformat{section}{\S\S#3#1#4 to~#5#2#6}
\crefmultiformat{section}{\S\S#2#1#3}{ and~#2#1#3}{, #2#1#3}{ and~#2#1#3}

\newcommand{\muhao}[1]{{\color{blue}MC:\textbf{#1}}}

\newcommand{\blue}[1]{\textcolor{blue}{#1}}

\definecolor{kaifu}{RGB}{65,50,132}

\aclfinalcopy 

\newcommand{\dr}[1]{\textcolor{blue}{[DR: #1]}}
\newcommand{\drc}[1]{\textcolor{blue}{#1}}

\usepackage{xcolor}

\newcommand{\ignore}[1]{}

\title{Joint Constrained Learning for Event-Event Relation Extraction}

\author{Haoyu Wang$^1$, Muhao Chen$^{1}$, Hongming Zhang$^2$\thanks{\indent This work was done when the author was visiting the University of Pennsylvania.}\; \& Dan Roth$^1$\\
$^1$Department of Computer and Information Science, UPenn\\
$^2$Department of Computer Science and Engineering, HKUST\\
\texttt{\{why16gzl, muhao, danroth\}@seas.upenn.edu};  \texttt{hzhangal@cse.ust.hk}\\ 
}

\date{}

\begin{document}
\maketitle

\begin{abstract}
   Understanding natural language involves recognizing how multiple event mentions structurally and temporally interact with each other. 
   In this process, one can induce event complexes that organize multi-granular events with temporal order and membership relations interweaving among them.
   Due to the lack of jointly labeled data for these relational phenomena and the restriction on the structures they articulate, we propose a joint constrained learning framework for modeling event-event relations.
   Specifically, the framework enforces logical constraints within and across multiple temporal and subevent relations 
   by converting these constraints into differentiable learning objectives. We show that our joint constrained learning approach effectively compensates for the lack of jointly labeled data, and outperforms SOTA methods on benchmarks for both temporal relation extraction and event hierarchy construction, replacing a commonly used but more expensive global inference process.
   We also present a promising case study showing the effectiveness of our approach in inducing event complexes on an external corpus.\footnote{Our code is publicly available at \url{https://cogcomp.seas.upenn.edu/page/publication_view/914}.} 

\end{abstract}



\section{Introduction}

\begin{figure}[t]
\begin{minipage}{\linewidth}
\noindent
\fbox{%
    \parbox{0.98\linewidth}{
    \fontsize{11pt}{13pt}\selectfont
    On Tuesday, there was a typhoon-strength (\textbf{\textit{e$_1$:storm}}) in Japan. One man got (\textbf{\textit{e$_2$:killed}}) and thousands of people were left stranded. Police said an 81-year-old man (\textbf{\textit{e$_3$:died}}) in central Toyama when the wind blew over a shed, trapping him underneath. Later this afternoon, with the agency warning of possible tornadoes, Japan Airlines (\textbf{\textit{e$_4$:canceled}}) 230 domestic flights, (\textbf{\textit{e$_5$:affecting}}) 31,600 passengers.

    }%
}

\vspace{0.2em}

    \centering
    \includegraphics[width=\textwidth]{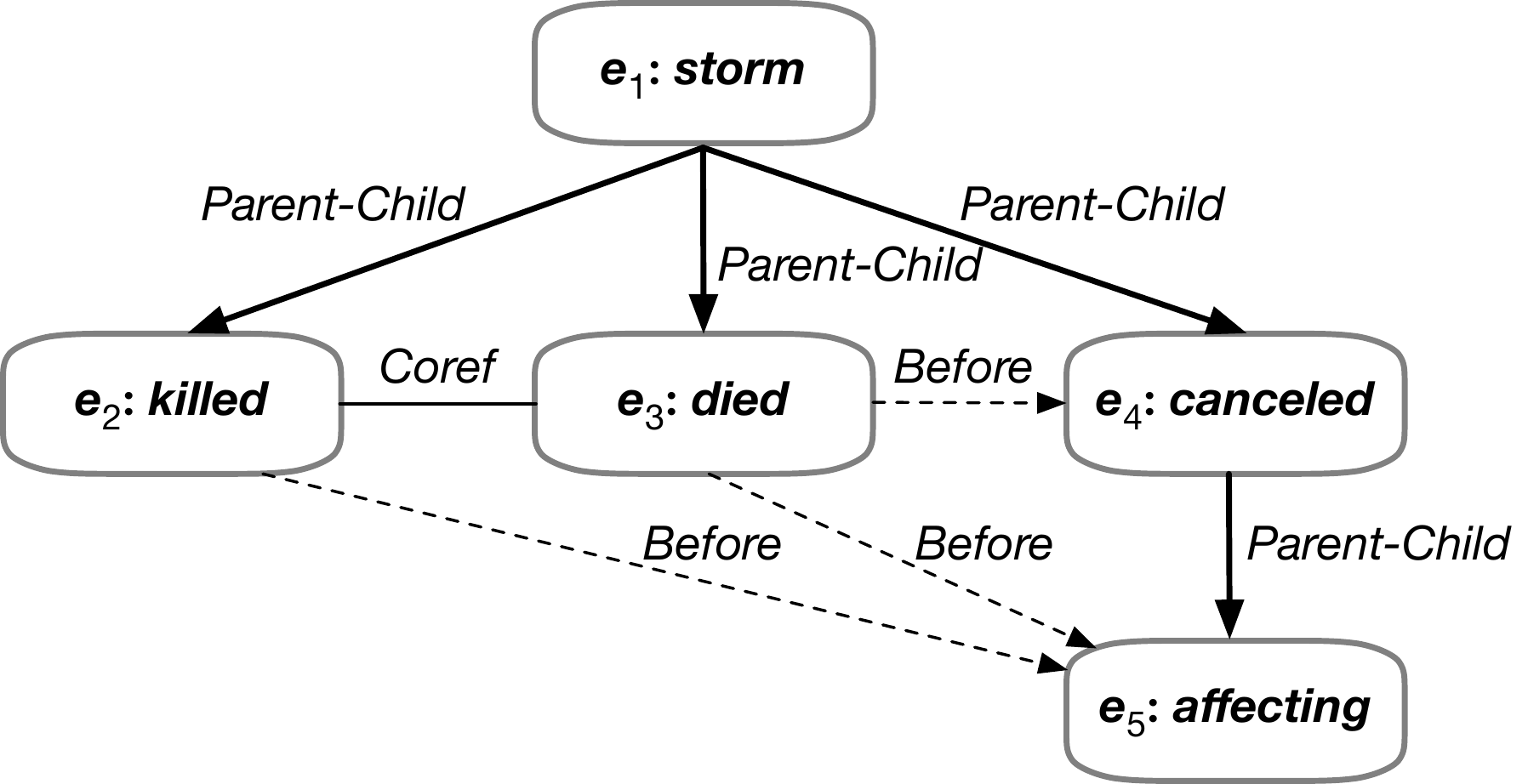}
    
    \caption{An example of an event complex described in the document. Bold arrows denote \textsc{Parent-Child} relation; dotted arrows represent \textsc{Before} relation; solid line represents two events are \textsc{Coref} to each other. For clarity, not all event mentions are shown in the figure.}
    \label{fig:example}
\end{minipage}
\end{figure}

Human languages evolve to communicate about 
real-world events. Therefore, understanding events plays a critical role in natural language understanding (NLU).
A key challenge to this mission lies in the fact that events are not just simple, standalone predicates.
Rather, they are often described at different granularities and may form complex structures. 
Consider the example in Figure~\ref{fig:example},
where the description of a storm (\emph{\textbf{e}}$_1$) involves more fine-grained event mentions about people killed (\emph{\textbf{e}}$_2$), flights canceled (\emph{\textbf{e}}$_3$) and passengers affected (\emph{\textbf{e}}$_4$).
Some of those mentions also follow strict temporal order (\emph{\textbf{e}}$_3$, \emph{\textbf{e}}$_4$ and \emph{\textbf{e}}$_5$).
Our goal is to induce such an \emph{event complex} that recognizes 
the membership of multi-granular events described in the text, as well as their temporal order.
This is not only at the core of text understanding,
but is also beneficial to various applications such as question answering \cite{khashabi2018question}, narrative prediction \cite{chaturvedi2017story}, timeline construction \cite{do-etal-2012-joint} and summarization \cite{daume-iii-marcu-2006-bayesian}.

Recently, significant 
research effort has been devoted to several event-event relation extraction tasks, such as event temporal relation (TempRel) extraction \cite{ning-etal-2018-improving,ning-etal-2019-improved} and subevent relation extraction \cite{liu-etal-2018-graph,aldawsari-finlayson-2019-detecting}.
Addressing such challenging tasks requires a model to recognize the inherent connection between event 
mentions as well as their contexts in the documents.
Accordingly, a few previous methods apply statistical learning methods to characterize the grounded events in the documents \cite{glavas-etal-2014-hieve, ning-etal-2017-structured,ning-etal-2018-cogcomptime}.
Such methods often require designing various features to characterize the structural, discourse and narrative aspects of the events, which are costly to produce and are often specific to a certain task or dataset.
More recent works attempted to use data-driven methods based on neural relation extraction models \cite{dligach-etal-2017-neural,ning-etal-2019-improved,han-etal-2019-deep,han-etal-2019-joint}
which refrain from feature engineering and offer competent performances.

While data-driven methods provide a
general and tractable way for event-event relation extraction, their performance is restricted by the limited annotated resources available~\cite{glavas-etal-2014-hieve,ning-etal-2018-multi}.
For example, the largest temporal relation extraction dataset MATRES~\cite{ning-etal-2018-multi} only has 275 articles, which is far from enough for training a well-performing supervised model.
The observation that relations and, in particular, event-event relations should be constrained by their logical properties \cite{RothYi04, chambers-jurafsky-2008-jointly}, led to employing global inference to comply with transitivity and symmetry consistency, specifically on TempRel \cite{DoLuRo12,ning-etal-2017-structured,han-etal-2019-deep}.
However, in an event complex, the logical constraints may globally apply to different task-specific relations, and form more complex conjunctive constraints. 
Consider the example in
Figure \ref{fig:example}: given that \textbf{\textit{e2:died}} is \textsc{Before} \textbf{\textit{e3:canceled}} and \textbf{\textit{e3:canceled}} is a \textsc{Parent} event of \textbf{\textit{e4:affecting}}, the learning process should enforce \textbf{\textit{e2:died}} \textsc{Before} \textbf{\textit{e4:affecting}} by considering the conjunctive constraints on both TempRel and subevent relations.
While previous works focus on preserving logical consistency through (post-learning) inference or structured learning \cite{NingFeRo17}, there was no 
effective way to endow neural models with the sense of global logical consistency during training. 
This is key to bridging 
the learning processes of 
TempRel and subevent relations, which is a research focus of this paper.


The \emph{first} contribution of this work is proposing 
a joint constrained learning model for multifaceted event-event relation extraction. 
The joint constrained learning framework seeks to regularize the model towards consistency with the logical constraints across both temporal and subevent relations, for which three types of consistency requirements are considered: \emph{annotation consistency}, \emph{symmetry consistency} and \emph{conjunction consistency}.
Such consistency requirements comprehensively define the interdependencies among those relations, essentially unifying the ordered nature of time and the topological nature of multi-granular subevents based on a set of declarative logic rules.
Motivated by the logic-driven framework proposed by \citet{li-etal-2019-logic},
the declarative logical constraints are converted into differentiable functions that can be incorporated into the learning objective for relation extraction tasks. 
Enforcing logical constraints across temporal and subevent relations is also a natural way to combine 
the supervision signals coming from two different datasets, one for each of the  relation extraction tasks with a shared learning objective. 
Despite the scarce annotation for both tasks, the proposed method surpasses the SOTA TempRel extraction method on MATRES by relatively 3.27\% in $F_1$;
it also offers promising performance on the HiEve dataset for subevent relation extraction, relatively surpassing previous methods by at least 3.12\% in $F_1$. 

From the NLU perspective, 
the \emph{second} contribution of this work lies in providing a general method for inducing an event complex that comprehensively represents the relational structure of several related event
mentions.
This is supported by the memberships vertically identified between multi-granular events, as well as the horizontal temporal reasoning within the event complex.
As far as we know, this is 
different from all 
previous works that only formulated relations along a single axis.
Our model further demonstrates the potent capability of inducing event complexes 
when evaluated 
on the RED dataset \cite{ogorman-etal-2016-richer}.

\section{Related Work}

Various approaches have been proposed to extract event TempRels.
Early effort focused on characterizing event pairs based on various types of semantic and linguistic features, and utilizing statistical learning methods, such as logistic regression  \cite{mani-etal-2006-machine, verhagen-pustejovsky-2008-temporal} and SVM \cite{mirza-tonelli-2014-classifying}, to capture the relations.
Those methods typically require extensive feature engineering, and do not comprehensively consider the contextual information and global constraints among event-event relations. 
Recently, data-driven methods have been developed for TempRel extraction, and have offered promising performance. 
\citet{ning-etal-2019-improved} addressed this problem using a system combining an LSTM document encoder and a Siamese multi-layer perceptron (MLP) encoder for temporal commonsense knowledge from \textsc{TemProb} \cite{ning-etal-2018-improving}. \citet{han-etal-2019-deep} proposed a bidirectional LSTM (BiLSTM) with structured prediction to extract TempRels.
Both of these works incorporated global inference to facilitate constraints on TempRels.


Besides TempRels, a couple of efforts have focused on event hierarchy construction, a.k.a. subevent relation extraction. This task seeks to extract the hierarchy where each parent event contains child events that are described in the same document.
To cope with this task, both \citet{araki-etal-2014-detecting} and \citet{glavas-snajder-2014-constructing} introduced a variety of features and employed logistic regression models for classifying event pairs into subevent relations (\textsc{Parent-Child} and \textsc{Child-Parent}, coreference (\textsc{Coref}), and no relation (\textsc{NoRel}).  
\citet{aldawsari-finlayson-2019-detecting} further extended the characterization with more features on the discourse and narrative aspects.
\citet{ZNKR20} presented a data-driven method by fine-tuning a time duration-aware BERT \cite{devlin-etal-2019-bert} on corpora of time mentions,
and used the estimation of time duration to predict subevent relations.


Though previous efforts have been devoted to preserving logical consistency through inference or structured learning~\cite{RothYi04,RothYi07,CRRR08}, this is difficult to do in the context of neural networks.
Moreover, while it is a common strategy to combine multiple training data in multi-task learning \cite{lin-ji-2020-joint}, our work is distinguished by enhancing the learning process by pushing the model towards a coherent output that satisfies logical constraints across separate tasks.

\section{Methods} 



In this section, we present the joint learning framework for event-event relation extraction. We start with the problem formulation (\Cref{sec:prelim}), followed by the techniques for event pair characterization (\Cref{sec:encoder}), constrained learning (\Cref{subsec:learningObjective}) and inference (\Cref{sec:inference}).

\begin{figure*}
    \centering
    \includegraphics[scale=0.5]{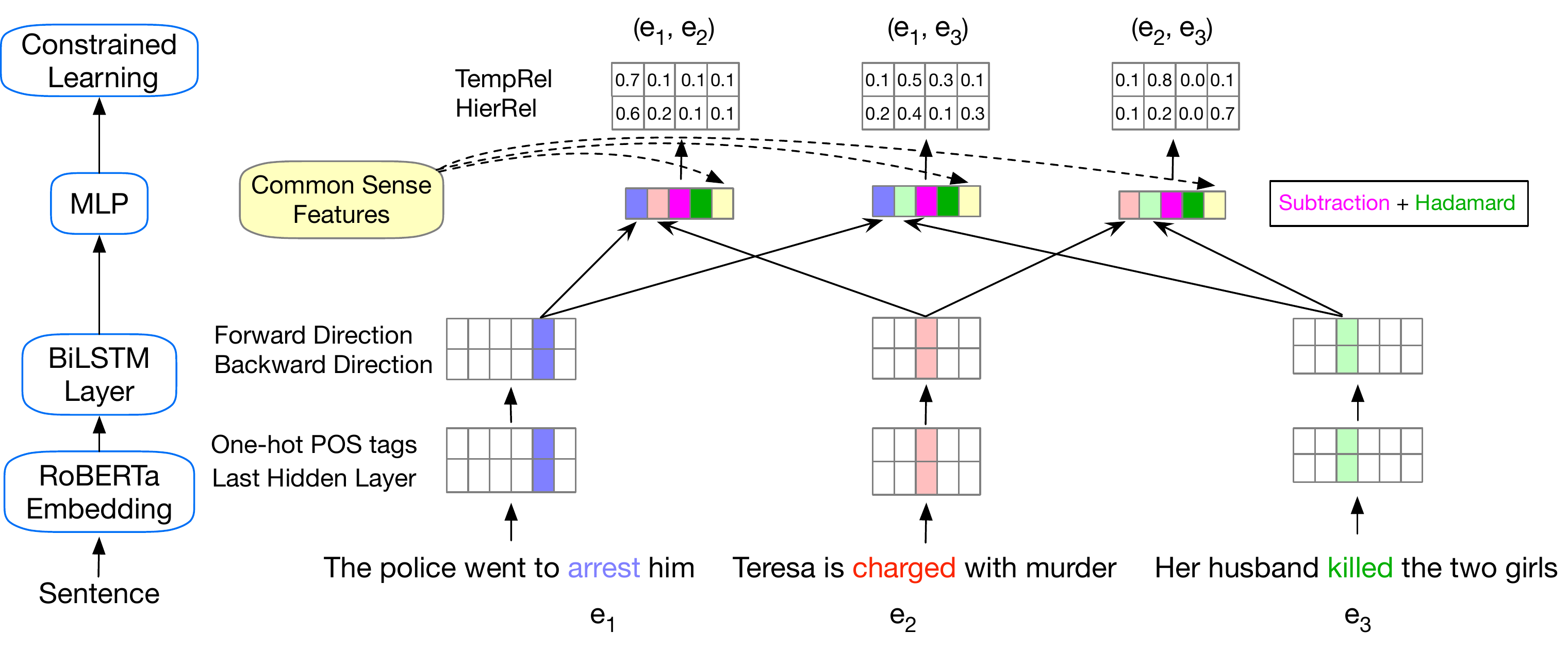}
    \caption{Model architecture. The model incorporates contextual features and commonsense knowledge to represent event pairs (\Cref{sec:encoder}). The joint learning enforces logical consistency on TempRel and subevent relations (\Cref{subsec:learningObjective}).
    }
    \label{fig:pipeline}
\end{figure*}

\subsection{Preliminaries}\label{sec:prelim}
A document $D$ is represented as a sequence of tokens $D = [t_1, \cdots, e_1, \cdots, e_2, \cdots, t_n]$.
Some of the tokens belong to the set of annotated event triggers, i.e., $\mathcal{E}_D = \{e_1, e_2, \cdots, e_k\}$,
whereas the rest are other lexemes.
The goal is to induce event complexes from the document, which is through extracting the multi-faceted event-event relations.
Particularly, we are interested in two subtasks of relation extraction, corresponding to the label set
$\mathcal{R} = \mathcal{R}_T \cup \mathcal{R}_H$.
$\mathcal{R}_T$ thereof denotes the set of temporal relations defined in the literature \cite{ning-etal-2017-structured,ning-etal-2018-multi,ning-etal-2019-improved,han-etal-2019-joint}, which contains \textsc{Before, After, Equal}, and \textsc{Vague}. 
To be consistent with previous studies \cite{ning-etal-2018-multi,ning-etal-2019-improved}, 
the temporal ordering relations between two events are decided by the order of their starting time, without constraining on their ending time.
$\mathcal{R}_H$ thereof denotes the set of relation labels defined in the subevent relation extraction task \cite{hovy-etal-2013-events,glavas-etal-2014-hieve}, i.e., \textsc{Parent-Child, Child-Parent}, \textsc{Coref} and \textsc{NoRel}. 
Following the definitions by \citet{hovy-etal-2013-events}, an event $e_1$ is said to have a child event $e_2$ if $e_1$ is a collector event that contains a sequence of activities, where $e_2$ is one of these activities, and $e_2$ is spatially and temporally contained within $e_1$.
Note that each pair of events can be annotated with one relation from each of $\mathcal{R}_H$ and $\mathcal{R}_T$ respectively, as the labels within each task-specific relation set are mutually exclusive.


Our learning framework first obtains the event pair representation that combines contextualized and syntactic features along with commonsense knowledge, and then use an MLP to get confidence scores for each relation in $\mathcal{R}$.
The joint learning objective seeks to enforce the logical consistency of outputs for both TempRel and subevent relations.
The overall architecture is shown in Figure \ref{fig:pipeline}.

\subsection{Event Pair Representation}\label{sec:encoder}

To characterize the event pairs in the document, we employ a neural encoder architecture which provides event representations from two groups of features.
Specifically, the representation here incorporates the contextualized representations of the event triggers along with statistical commonsense knowledge from several knowledge bases. On top of the features that characterize an event pair $(e_1, e_2)$, we use an MLP with $|\mathcal{R}|$ outputs to estimate the confidence score for each relation $r$, denoted as $r_{(e_1, e_2)}$.
Two separate softmax functions are then added to normalize the outputs for two task-specific label sets $\mathcal{R}_T$ and $\mathcal{R}_H$.

\subsubsection{Contextualized Event Trigger Encoding}\label{sec:roberta}

Given a document,
we first use a pre-trained language model, RoBERTa \cite{liu2019roberta}, to produce the contextualized embeddings for all tokens of the entire document. 
The token embeddings are further concatenated with the one-hot vectors of POS (part-of-speech) tags, and fed into a BiLSTM. 
The hidden state of the last BiLSTM layer that is stacked on top of each event trigger $e$ is therefore treated as the embedding representation of the event, denoted as $h_{e}$. 
For each event pair $(e_1, e_2)$, the contextualized features are obtained as the concatenation of $h_{e_1}$ and $h_{e_2}$, along with their element-wise Hadamard product and subtraction. This is shown to be a comprehensive way to model embedding interactions \cite{ZCJWJW20}.

\subsubsection{Commonsense Knowledge}\label{sub:cse}

We also incorporate the following sources of commonsense knowledge to characterize event pairs.
Specifically, we first extract relevant knowledge from ConceptNet \cite{speer2017conceptnet},
which is a large-scale commonsense knowledge graph for commonsense concepts, entities, events and relations.
A portion of the relations in ConceptNet that are relevant to our tasks include ``HasSubevent",
``HasFirstSubevent" and ``HasLastSubevent" relations.
From ConceptNet we extract around 30k pairs of event concepts labeled with the aforementioned relations, along with 30k randomly corrupted negative samples. We also incorporate commonsense knowledge from \textsc{TemProb} \cite{ning-etal-2018-improving}.
This provides prior knowledge of the temporal order that some events usually follow.

We use the event pairs from those knowledge bases to train two MLP encoders.
Each takes the concatenated token embeddings of two event triggers as inputs, and is trained with \emph{contrastive loss} to estimate the likelihood that if a relation holds.
For subevent and temporal related commonsense knowledge, two MLPs are separately trained.
After the encoders are well-trained, we fix their parameters and combine them as a black box that corresponds to ``Common Sense Features" in Figure \ref{fig:pipeline}.
\begin{table*}[]
    \centering
    \small
    \setlength{\tabcolsep}{0.5pt}
    \begin{tabular}{|c|c|c|c|c|c|c|c|c|}\hline
     \backslashbox{$\alpha$}{$\beta$} & PC & CP & CR & NR& \blue{BF} & \blue{AF} & \blue{EQ} & \blue{VG} \\\hline
     PC & PC, $\neg$\blue{AF} & -- & PC, $\neg$\blue{AF}  & $\neg$CP, $\neg$CR & \blue{BF} , $\neg$CP, $\neg$CR & -- & \blue{BF} , $\neg$CP, $\neg$CR & -- \\\hline
     CP & -- & CP, $\neg$\blue{BF} & CP, $\neg$\blue{BF} & $\neg$PC, $\neg$CR & -- & \blue{AF}, $\neg$PC, $\neg$CR & \blue{AF}, $\neg$PC, $\neg$CR & -- \\\hline
     CR & PC, $\neg$\blue{AF} & CP, $\neg$\blue{BF} & CR, \blue{EQ} & NR& \blue{BF} , $\neg$CP, $\neg$CR & \blue{AF}, $\neg$PC, $\neg$CR & \blue{EQ} & \blue{VG} \\\hline
     NR& $\neg$CP, $\neg$CR & $\neg$PC, $\neg$CR & NR& -- & -- & -- & -- & -- \\\hline
     \blue{BF} & \cellcolor{orange!25}\blue{BF} , $\neg$CP, $\neg$CR & -- & \blue{BF} , $\neg$CP, $\neg$CR & -- & \blue{BF} , $\neg$CP, $\neg$CR & -- & \blue{BF} , $\neg$CP, $\neg$CR & $\neg$\blue{AF}, $\neg$\blue{EQ} \\\hline
     \blue{AF} & -- & \blue{AF}, $\neg$PC, $\neg$CR & \blue{AF}, $\neg$PC, $\neg$CR & -- & -- & \blue{AF}, $\neg$PC, $\neg$CR & \blue{AF}, $\neg$PC, $\neg$CR & $\neg$\blue{BF} , $\neg$\blue{EQ} \\\hline
     \blue{EQ} & $\neg$\blue{AF} & $\neg$\blue{BF} & \blue{EQ} & -- & \blue{BF} , $\neg$CP, $\neg$CR & \blue{AF}, $\neg$PC, $\neg$CR & \blue{EQ} & \blue{VG}, $\neg$CR\\\hline
     \blue{VG} & -- & -- & \blue{VG}, $\neg$CR & -- & $\neg$\blue{AF}, $\neg$\blue{EQ} & $\neg$\blue{BF} , $\neg$\blue{EQ} & \blue{VG} & -- \\\hline
    \end{tabular}
    \caption{The induction table for conjunctive constraints on temporal and subevent relations. 
    Given the relations $\alpha(e_1,e_2)$ in the left-most column and $\beta(e_2,e_3)$ in the top row,
    each entry in the table includes all the relations and negations that can be deduced from their conjunction for $e_1$ and $e_3$, i.e. De$(\alpha, \beta)$. The abbreviations PC, CP, CR, NR, BF, AF, EQ and VG denote \textsc{Parent-Child}, \textsc{Child-Parent}, \textsc{Coref}, \textsc{NoRel}, \textsc{Before}, \textsc{After}, \textsc{Equal} and \textsc{Vague}, respectively.
    Vertical relations are in black, and TempRel are in blue. ``--" denotes no constraints. }
    \label{tab:constraints}
\end{table*}
\subsection{Joint Constrained Learning}
\label{subsec:learningObjective}


Given the characterization of grounded event pairs from the document, we now define the learning objectives for relation prediction.
The goal of learning is to let the model capture the data annotation, meanwhile regularizing the model towards consistency on logic constraints.
Inspired by the logic-driven framework for consistency of neural models \cite{li-etal-2019-logic}, we specify three types of consistency requirements, i.e. \emph{annotation consistency}, \emph{symmetry consistency} and \emph{conjunction consistency}.
We hereby define the requirements with declarative logic rules, and show how we transform them into differentiable loss functions.

\paragraph{Annotation Consistency}
For labeled cases, we expect the model to predict what annotations specify. That is to say, if an event pair is annotated with relation $r$, then the model should predict so: 
\begin{align*}
  \Land_{e_1, e_2 \in \mathcal{E}_D} \top \rightarrow r(e_1, e_2).
\end{align*}
\noindent 
To obtain the learning objective that preserves the annotation consistency,
we use the product t-norm to get the learning objective of maximizing the probability of the true labels, by transforming to the negative log space to capture the inconsistency with the product t-norm. Accordingly, the
annotation loss is equivalently defined as the cross entropy
\begin{align*}
  L_{A} = \sum_{e_1, e_2\in \mathcal{E}_D} - w_r \log r_{\p{e_1, e_2}},
\end{align*}
\noindent
in which $w_r$ is the label weight that seeks to balance the loss for training cases of each relation $r$.
\paragraph{Symmetry Consistency}
Given any event pair $(e_1, e_2)$, the grounds for a model to predict a relation $\alpha(e_1, e_2)$ to hold between them should also implies the hold of the converse relation $\bar{\alpha}(e_2, e_1)$. 
The logical formula is accordingly written as
\begin{align*}
\Land_{e_1, e_2 \in \mathcal{E}_D, \; \alpha \in \mathcal{R}_S} \alpha(e_1, e_2) \leftrightarrow \bar{\alpha}(e_2, e_1),
\label{logic:sym}
\end{align*}
\noindent
where the $\mathcal{R}_S$ is the set of relations enforcing the symmetry constraint. 
Particularly for the TempRel extraction task, $\mathcal{R}_S$ contains a pair of reciprocal relations \textsc{Before} and \textsc{After}, as well as two reflexive ones \textsc{Equal} and \textsc{Vague}.
Similarly, the subevent relation extraction task adds reciprocal relations \textsc{Parent-Child} and \textsc{Child-Parent} as well as reflexive ones \textsc{Coref} and \textsc{NoRel}.

Using the product t-norm and transformation to the negative log space as before, we have the symmetry loss:
\begin{align*}
  L_{S} =  \sum_{e_1, e_2\in \mathcal{E}, \alpha \in \mathcal{R}_S} | \log \alpha_{\p{e_1, e_2}} - & \log \bar{\alpha}_{(e_2, e_1)} | .
\end{align*}
\paragraph{Conjunction Consistency}
This set of constraints are applicable to any three related events $e_1, e_2$ and $e_3$. If we group the events into three pairs, namely $(e_1, e_2), (e_2, e_3)$ and $(e_1, e_3)$,  the relation definitions mandate that not all of the possible assignments to these three pairs are allowed.
More specifically, if two relations $\alpha{(e_1, e_2)}$ and $\beta{(e_2, e_3)}$ apply to the first two pairs of events,
then the conjunction consistency may enforce the following two conjunctive rules. 

In the first rule, the conjunction of the first two relations infers the hold of another relation $\gamma$ between the third event pair $(e_1, e_3)$, namely
\begingroup\makeatletter\def\f@size{10}\check@mathfonts
\begin{align*}
\begin{split}
    \Land_{\substack{e_1,e_2,e_3 \in \mathcal{E}_D\\ \alpha, \beta \in \mathcal{R}, \text{ }\gamma \in \text{De}(\alpha, \beta)}} 
    \alpha(e_1, e_2) \wedge \beta(e_2, e_3) \rightarrow	\gamma(e_1, e_3) .
\end{split}
\end{align*}
\endgroup
\noindent $\text{De}(\alpha, \beta)$ thereof is a set composed of all relations from $\mathcal{R}$ that do not conflict with $\alpha$ and $\beta$,
which is a subset of the deductive closure \cite{stine1976skepticism} of the conjunctive clause for these two relations.
A special case that the above formula expresses is a (task-specific) transitivity constraint, where $\alpha=\beta=\gamma$ present the same transitive relation.

Another condition could also hold, where the former two relations always infer the negation of a certain relation $\delta$ on $(e_1,e_3)$, for which we have
\begingroup\makeatletter\def\f@size{10}\check@mathfonts
\begin{align*}
\begin{split}
    \Land_{\substack{e_1,e_2,e_3 \in \mathcal{E}_D\\ \alpha, \beta \in \mathcal{R}, \text{ }\delta \notin \text{De}(\alpha, \beta)}} \alpha(e_1, e_2) \wedge \beta(e_2, e_3) \rightarrow	\neg\delta(e_1, e_3) .
\end{split}
\end{align*}
\endgroup
\noindent
\Cref{tab:constraints} is an induction table that describes all the conjunctive rules for relations in $\mathcal{R}$. To illustrate the conjunction consistency requirement (see the orange cell in \Cref{tab:constraints}), assume that $(e_1,e_2)$ and $(e_2,e_3)$ are respectively annotated with \textsc{Before} and \textsc{Parent-Child}.
Then the two conjunctive formulae defined above
infer that we have the relation \textsc{Before} hold on $(e_1,e_3)$, whereas we should not have \textsc{Child-Parent} hold.

Similar to the other consistency requirements, the loss function dedicated to the conjunction consistency is derived as follows:
\begin{align*}
\begin{split}
    L_{C} & = \sum_{\substack{e_1, e_2, e_3\in \mathcal{E}_D, \\\alpha, \beta \in \mathcal{R}, \gamma \in \text{De}(\alpha, \beta)}} |L_{t_1}|
    + \sum_{\substack{e_1, e_2, e_3\in \mathcal{E}_D, \\\alpha, \beta \in \mathcal{R}, \delta \notin \text{De}(\alpha, \beta)}} |L_{t_2}| ,
\end{split}
\end{align*}
\noindent
where the two terms of triple losses are defined as
\begin{align*}
\begin{split}
    L_{t_1} =& \log \alpha_{\p{e_1, e_2}} +  \log \beta_{\p{e_2, e_3}} - \log \gamma_{\p{e_1, e_3}}\\
    L_{t_2} =&  \log \alpha_{\p{e_1, e_2}} +  \log \beta_{\p{e_2, e_3}} - \log (1 - \delta_{\p{e_1, e_3}})
\end{split}
\end{align*}
\noindent
It is noteworthy that modeling the conjunctive consistency is key to the combination of two different event-event relation extraction tasks, 
as this general consistency requirement can be enforced between both TempRels and subevent relations.

\paragraph{Joint Learning Objective}
After expressing the logical consistency requirements with different terms of cross-entropy loss,
we combine all of those into the following joint learning objective loss
\begin{align*}
  L = L_{A} + \lambda_{S} L_{S} + \lambda_{C} L_{C} .
\end{align*}
The $\lambda$'s are non-negative coefficients to control the influence of each loss term.
Note that since the consistency requirements are defined on both temporal and subevent relations,
the model therefore seamlessly incorporates both event-event relation extraction tasks with a shared learning objective.
In this case, the learning process seeks to unify the ordered nature of time and the topological nature of subevents,
therefore supporting the model to comprehensively understand the event complex.

\subsection{Inference}\label{sec:inference}

To support task-specific relation extraction, i.e. extracting either a TempRel or a subevent relation, our framework selects the relation $r$ with highest confident score $r_{(e_1,e_2)}$ from either of $\mathcal{R}_T$ and $\mathcal{R}_H$.
When it comes to extracting event complexes with both types of relations, the prediction of subevent relations has higher priority. The reason lies in the fact that a relation in $\mathcal{R}_H$, except for \textsc{NoRel}, always implies a TempRel, yet there is not a single TempRel that necessitates a subevent relation.

We also incorporate ILP in the inference phase to further ensure the logical consistency in predicted results. Nevertheless, we show in experiments that a well-trained constrained learning model may not additionally require global inference (\Cref{sec:ablation}).

 
\section{Experiments}
In this section, we present the experiments on event-event relation extraction.
Specifically, we conduct evaluation for TempRel and subevent relation extraction based on two benchmark datasets (\Cref{sec:dataset}-\Cref{sec:results}).
To help understand the significance of each model component in the framework, we also give a detailed ablation study (\Cref{sec:ablation}).
Finally, a case study on the RED dataset is described to demonstrate the capability of inducing event complexes (\Cref{sec:red}). 

\subsection{Datasets}\label{sec:dataset}
Since there is not a large-scale dataset that amply annotates for both TempRel and subevent relations,
we evaluate the joint training and prediction of both categories of relations on two separate datasets.
Specifically, we use MATRES \cite{ning-etal-2018-multi} for TempRel extraction and HiEve \cite{glavas-etal-2014-hieve} for subevent relation extraction.

MATRES is a new benchmark dataset for TempRel extraction, which is developed from TempEval3 \cite{uzzaman-etal-2013-semeval}.
It annotates on top of 275 documents with TempRels \textsc{Before, After, Equal}, and \textsc{Vague}. 
Particularly, the annotation process of MATRES has defined four axes for the actions of events, i.e. \emph{main}, \emph{intention}, \emph{opinion}, and \emph{hypothetical} axes.
The TempRels are considered for all event pairs on the same axis and within a context of two adjacent sentences.
The labels are decided by comparing the starting points of the events.
The multi-axis annotation helped MATRES to achieve a high IAA of 0.84 in Cohen’s Kappa.

The HiEve corpus is a news corpus that contains 100 articles.
Within each article, annotations are given for both subevent and coreference relations.
The HiEve adopted the IAA measurement proposed for TempRels by \cite{uzzaman-allen-2011-temporal}, resulting in 0.69 $F_1$. 

In addition to these two datasets, we also present a case study on an updated version of the RED dataset \cite{ogorman-etal-2016-richer}. 
This dataset contains 35 news articles with annotations for event complexes that contain both membership relations and TempRels.
Since small dataset is not sufficient for training,
we use it only to demonstrate our method's capability of inducing event complexes on data that are external to training. 

We briefly summarize the data statistics for HiEve, MATRES, and RED dataset in Table~\ref{tbl:stats}. 


\subsection{Baselines and Evaluation Protocols}
On MATRES, we compare with four baseline methods. 
\citet{ning-etal-2018-multi} present a baseline method based on a set of linguistic features and an averaged perceptron classifier (Perceptron).
\citet{han-etal-2019-joint} introduce a BiLSTM model that incorporates MAP inference (BiLSTM+MAP). \citet{ning-etal-2019-improved} present the SOTA data-driven method incorporating ILP and  commonsense knowledge from \textsc{TemProb} with LSTM (LSTM+CSE+ILP).
We also compare with the
CogCompTime system \cite{ning-etal-2018-cogcomptime}.
On HiEve\footnote{Despite carefully following the details described in \cite{aldawsari-finlayson-2019-detecting} and communicating with the authors, we were not able to reproduce their results. Therefore, we choose to compare with other methods.}, 
we compare with a structured logistic regression model (StructLR, \citealt{glavas-snajder-2014-constructing}) and a recent data-driven method based on fined-tuning a time duration-aware BERT on large time-related web corpora (\textsc{TacoLM}, \citealt{ZNKR20}).

\begin{table}[!t]
    \centering
    \setlength{\tabcolsep}{3pt}
    {%
    \small
    \begin{tabular}{l|ccc}
    \hline
    \toprule
     Model & $P$ & $R$ & $F_1$ \\ \hline
     CogCompTime \cite{ning-etal-2018-cogcomptime} & 0.616 & 0.725 &  0.666 \\
     Perceptron \cite{ning-etal-2018-multi}  & 0.660 & 0.723 & 0.690 \\
     BiLSTM+MAP \cite{han-etal-2019-joint} & - & - & 0.755 \\ 
     LSTM+CSE+ILP \cite{ning-etal-2019-improved} & 0.713 & 0.821 & 0.763 \\
       
    \CC{20}Joint Constrained Learning (ours) & \CC{20}\textbf{0.734} & \CC{20}\textbf{0.850} & \CC{20}\textbf{0.788}\\ 
    \bottomrule
    \end{tabular}
    }
    \caption{TempRel extraction results on MATRES. Precision and recall are not reported by \cite{han-etal-2019-joint}.}
    \label{tab:result_M}
\end{table}

\begin{table}[t]
\label{tab:stat}
    \centering
    \small
    \begin{tabular}{c|c|c|c}\toprule
         & HiEve & MATRES & RED\\\hline
         \multicolumn{4}{c}{\# of Documents}\\\hline
         Train & 80 & 183 & - \\
         Dev & - & 72 & - \\
         Test & 20 & 20 & 35\\
         \midrule
          \multicolumn{4}{c}{\# of Pairs}\\
          \midrule
       Train  & 35001 & 6332 & - \\
       Test  & 7093 & 827 & 1718\\
       \bottomrule
    \end{tabular}
    \caption{Data statistics of HiEve, MATRES, and RED.}\label{tbl:stats}
    
\end{table}

MATRES comes with splits of 183, 72 and 20 documents respectively used for training, development and testing.
Following the settings in previous work \cite{ning-etal-2019-improved,han-etal-2019-joint}, 
we report the micro-average of precision, recall and F1 scores on test cases.
On HiEve, we use the same evaluation setting as \citet{glavas-snajder-2014-constructing} and \citet{ZNKR20},
leaving 20\% of the documents out for testing. The results in terms of $F_1$ of \textsc{Parent-Child} and \textsc{Child-Parent} and the micro-average of them are reported. 
Note that in the previous setting by  \citet{glavas-snajder-2014-constructing},
the relations are only considered for event pairs $(e_1,e_2)$ where $e_1$ appears before $e_2$ in the document.
We also follow \citet{glavas-snajder-2014-constructing} to populate the annotations by computing the transitive closure of \textsc{CoRef} and subevent relations.


\subsection{Experimental Setup}

To encode the tokens of each document,
we employ the officially released 768 dimensional RoBERTa \cite{liu2019roberta}, which is concatenated with 18 dimensional one-hot vectors representing the tokens' POS tags.
On top of those embeddings, the hidden states of the trainable BiLSTM are 768 dimensional, and we only apply one layer of BiLSTM.
Since the TempRel extraction and subevent relation extraction tasks are considered with two separate sets of labels, we use two separate softmax functions for normalizing the outputs for each label set from the single MLP.
For all the MLPs we employ one hidden layer each, whose dimensionality is set to the average of the input and output space following convention \cite{chen-etal-2018-yeji}.

We use AMSGrad \cite{reddi2018convergence} to optimize the parameters, with the learning rate set to 0.001.
Label weights in the annotation loss $L_A$ is set to balance among training cases for different relations.
The coefficients $\lambda_S$ and $\lambda_D$ in the learning objective function are both fixed to $0.2$. 
Training is limited to 80 epochs, which is sufficient to converge.

\begin{table}[!t]
    \centering
    \setlength{\tabcolsep}{5pt}
    {
    \small
    \begin{tabular}{l|ccc}\hline
    \toprule
    & \multicolumn{3}{c}{$F_1$ score} \\
    Model & \textsc{PC} & \textsc{CP} & Avg. \\ \hline
    StructLR \cite{glavas-etal-2014-hieve}& 0.522 & \textbf{0.634} & 0.577 \\ 
    \textsc{TacoLM} \cite{ZNKR20} & 0.485 & 0.494 & 0.489 \\
    \CC{20}Joint Constrained Learning (ours)  & \CC{20}\textbf{0.625} & \CC{20}0.564 & \CC{20}\textbf{0.595} \\
    \bottomrule
    \end{tabular}
    }
    \caption{Subevent relation extraction results on HiEve. \textsc{PC}, \textsc{CP} and Avg. respectively denote \textsc{Parent-Child}, \textsc{Child-Parent} and their micro-average.}
    \label{tab:result_H}
\end{table}
\begin{table*}[!t]
    \centering
    {
    \small
    \begin{tabular}{l|ccc|ccc}\hline 
    \toprule
    & \multicolumn{3}{c|}{\textsc{Subevent}} &  \multicolumn{3}{c}{\textsc{TempRel}} \\
    Model & $P$ & $R$ & $F_1$ & $P$ & $R$ & $F_1$ \\ \hline
   Single-task Training & 32.5 & \textbf{73.1} & 45.0  & 67.7 & 80.3 & 73.5 \\
   
    Joint Training & 50.4 & 43.1 & 46.5 & 68.4 	& 82.0 & 74.6 \\
    \hline
    + Task-specific constrained learning & 51.6 & 59.7 & 55.4 & 71.3 & 82.7 & 76.6 \\
    + Cross-task constrained learning & 51.1 & 67.0 & 58.0 &  72.2 & 83.8 & 77.6\\  
    + Commonsense knowledge & 56.9 & 61.6 & 59.2  & 73.3 & 84.2 & 78.4\\
    \CC{20}+ Global inference (ILP) & \CC{20}\textbf{57.4} & \CC{20}61.7 & \CC{20}\textbf{59.5} & \CC{20}\textbf{73.4} & \CC{20}\textbf{85.0} & \CC{20}\textbf{78.8}  \\
    \midrule
    All but constrained learning & 54.2 & 41.8 & 47.2 & 72.1 & 80.8 & 76.2 \\
    \bottomrule
    \end{tabular}
    }
    \caption{Ablation study results (\Cref{sec:ablation}). The results on HiEve are the micro-average of \textsc{Parent-Child} and \textsc{Child-Parent}.
    Results in the middle group are achieved by incrementally adding the corresponding model components. The gray-scaled row shows the results of the complete model. 
    }
    \label{tab:Ablation}
\end{table*}

\subsection{Results}\label{sec:results}
In Table \ref{tab:result_M} we report the TempRel extraction results on MATRES.
Among the baseline methods, \citet{ning-etal-2019-improved} offer the best performance in terms of $F_1$ by incorporating an LSTM with global inference and commonsense knowledge. 
In contrast, the proposed joint constrained learning framework surpasses the best baseline method by a relative gain of 3.27\% in $F_1$, and excels in terms of both precision and recall.
While both methods ensure logical constraints in learning or inference phases, the improvement by the proposed method is largely due to the joint constraints combining both TempRel and subevent relations.
Learning to capture subevent relations from an extrinsic resource simultanously offer auxiliary supervision signals to improve the comprehension on TempRel, even though the resources dedicated to the later is limited.

The results in Table \ref{tab:result_H} for subevent relation extraction exhibit similar observation.
Due to scarcer annotated data, the pure data-driven baseline method (\textsc{TacoLM}) falls behind the statistical learning one (i.e. StructLR) with comprehensively designed features. 
However, our model successfully complements the insufficient supervision signals, partly by incorporating linguistic and commonsense knowledge.
More importantly, while our model is able to infer TempRel decently, the global consistency ensured by cross-task constraints naturally makes up for the originally weak supervision signals for subevent relations.
This fact leads to promising results, drastically surpassing \textsc{TacoLM} with a relative gain of 21.4\% in micro-average $F_1$,
and outperforming StructLR by $\sim$3\% relatively.

In general, the experiments here show that the proposed joint constrained learning approach effectively combines the scarce supervision signals for both tasks.
Understanding the event complex by unifying the ordered nature of time and the topological nature of multi-granular subevents, assists the comprehension on both TempRel and memberships among multi-granular events.

\subsection{Ablation Study}
\label{sec:ablation}
To help understand the model components,
we conduct an ablation study and report the results in Table \ref{tab:Ablation}.
Starting from the vanilla single-task BiLSTM model with only RoBERTa features, changing to joint training both tasks with only annotation brings along 1.1-1.5\% of absolute gain in $F_1$.
Incorporating task-specific constraints to learning for relations only in $\mathcal{R}_T$ or $\mathcal{R}_H$ notably brings up the $F_1$ 2.0-8.9\%, whereas the cross-task constraints bring along an improvement of 1.0-2.6\% in $F_1$.
This indicates that the global consistency ensured within and across TempRel and subevent relations is important for enhancing the comprehension for both categories of relations.
The commonsense knowledge leads to another 0.8-1.2\% of improvement.
Lastly, global inference does not contribute much to the performance in our setting, which indicates that the rest model components are already sufficient to preserve global consistency through joint constrained learning.

To compare both ways of ensuring logical consistency, we also report a set of results in the last row of Table~\ref{tab:Ablation}, where constrained learning is removed and only global inference is used to cope with consistency requirements in prediction.
As expected, this leads to significant performance drop of 2.6-12.3\% in $F_1$.
This fact implies that ensuring the logical consistency in the learning phase is essential, in terms of both complementing task-specific training and enhancing the comprehension of event complex components.


\subsection{Case Study on the RED Dataset}\label{sec:red}

\begin{table}[t]
\label{tab:mapping}
\small
\centering
\begin{tabular}{l|l}
\toprule
Original labels in RED & Mapped labels\\\hline
\begin{tabular}[c]{@{}l@{}}BEFORE,\\ BEFORE/CAUSES,\\ BEFORE/PRECONDITION,\\ ENDS-ON,\\ OVERLAP/PRECONDITION\end{tabular} & \textsc{Before}                                                         \\ \hline
SIMULTANEOUS                                                                                                              & \textsc{Equal}                                                          \\ \hline
\begin{tabular}[c]{@{}l@{}}OVERLAP,\\ REINITIATES\end{tabular}                                                            & \textsc{Vague}                                                          \\ \hline
\begin{tabular}[c]{@{}l@{}}CONTAINS,\\ CONTAINS-SUBEVENT\end{tabular}                                                     & \begin{tabular}[c]{@{}l@{}}\textsc{Parent-Child} \&\\ \textsc{Before}\end{tabular} \\ \hline
BEGINS-ON                                                                                                                 & \textsc{After}                                                          \\ \bottomrule
\end{tabular}
\caption{Mapping from relations annotated in the RED dataset to the relations studied in this work.}\label{tbl:RED_mapping}
\end{table}

We use the RED dataset (2019 updated version) to further evaluate our model trained on MATRES and HiEve for inducing complete event complexes,
as well as to show the model's generalizability to an external validation set.
Since the labels of RED are defined differently from those in the datasets we train the model on,
Table~\ref{tbl:RED_mapping} shows the details about how some RED lables are mapped to MATRES and HiEve labels.
Other event-event relations in RED are mapped to \textsc{Vague} or \textsc{NoRel} according to their relation types, and the relations annotated between entities are discarded.
To obtain the event complexes, 
as stated in \Cref{sec:inference}, prediction of subevent relations is given higher priority than that of TempRels.
In this way, our model achieves 0.72 $F_1$ on TempRel extraction and 0.54 $F_1$ on subevent relation extraction.

\begin{figure}[t]
\begin{minipage}{\linewidth}
\noindent
\fbox{%
    \parbox{0.98\linewidth}{
    \fontsize{11pt}{13pt}\selectfont
    A (\textbf{\textit{e1:convoy}}) of 280 Russian trucks (\textbf{\textit{e2:headed}}) for Ukraine, which Moscow says is (\textbf{\textit{e3:carrying}}) relief goods for war-weary civilians, has suddenly (\textbf{\textit{e4:changed}}) course, according to a Ukrainian state news agency.
    }%
}

\vspace{0.2em}

    \centering
    \includegraphics[width=\textwidth]{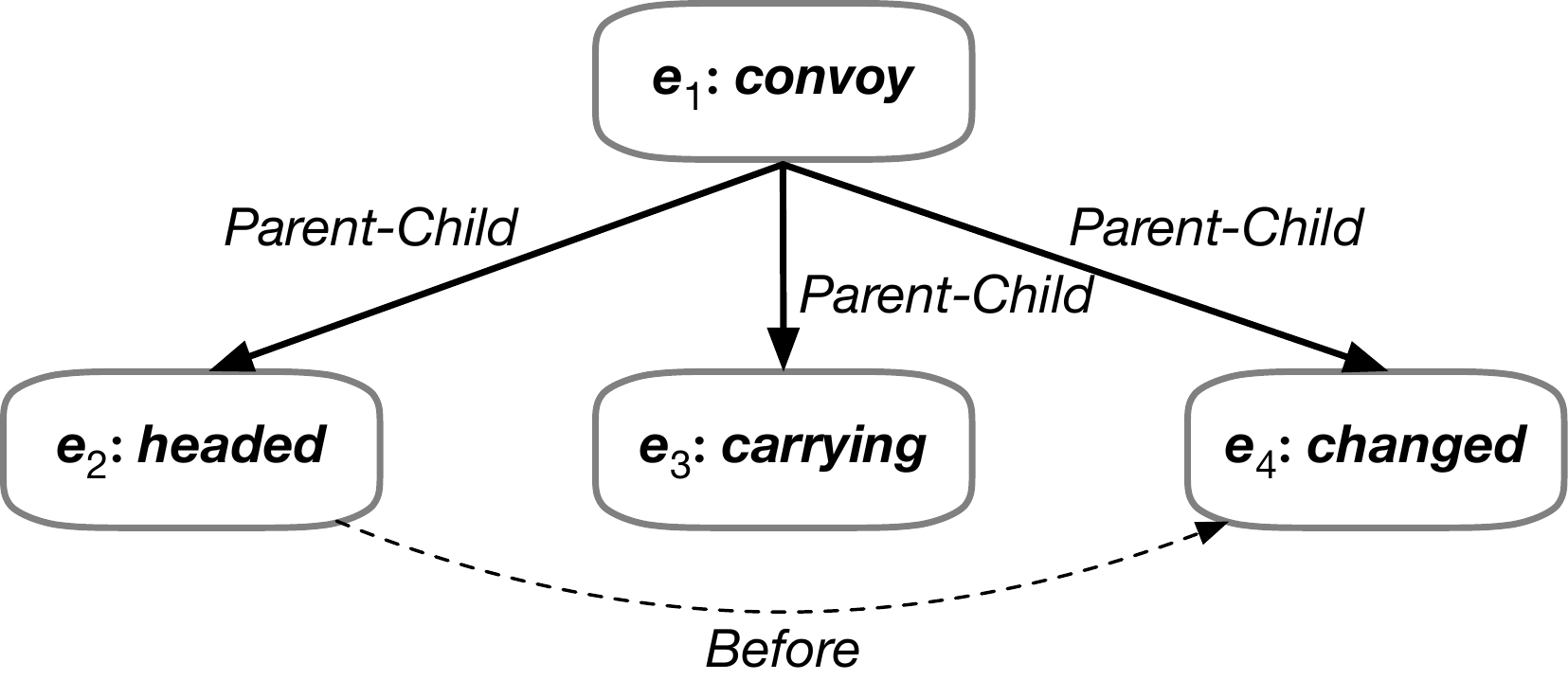}
    
    \caption{An example of an event complex extracted from a document in RED. Bold arrows denote the \textsc{Parent-Child} relation, and dotted arrows represent the \textsc{Before} relation.}
    \label{fig:RED_example}
\end{minipage}
\end{figure}

Here we give an example of an event complex extracted from the RED dataset in \Cref{fig:RED_example}, using our joint constrained learning method.

\section{Conclusion}
We propose a joint constrained learning framework for extracting event complexes from documents.
The proposed framework bridges TempRel and subevent relation extraction tasks with a comprehensive set of logical constraints, which are enforced during learning by  converting them into differentiable objective functions. 
On two benchmark datasets, the proposed method outperforms SOTA statistical learning methods and data-driven methods for each task, without using data that is jointly annotated with the two classes of relations. It also presents promising event complex extraction results on RED that is external to training. 
Thus, our work shows that the global consistency of the event complex significantly helps understanding both temporal order and event membership.
For future work, we plan to extend the framework towards an end-to-end system with event extraction. We also seek to extend the conjunctive constraints along with event argument relations.

\section*{Acknowledgement}

We appreciate the anonymous reviewers for their insightful comments.
Also, we would like thank Jennifer Sheffield and other members from the UPenn Cognitive Computation Group for giving suggestions to improve the manuscript.

This research is supported by the Oﬃce of the Director of National Intelligence (ODNI), Intelligence Advanced Research Projects Activity (IARPA), via IARPA Contract No. 2019-19051600006 under the BETTER Program, and by contract FA8750-19-2-1004 with the US Defense Advanced Research Projects Agency (DARPA). The views expressed are those of the authors and do not reflect the official policy or position of the Department of Defense or the U.S. Government.

\bibliography{anthology,emnlp2020,ccg.bib}
\bibliographystyle{acl_natbib}

\end{document}